# Thermal coupling and effect of subharmonic synchronization in a system of two VO$_2$ based oscillators

**Andrey Velichko\*, Maksim Belyaev, Vadim Putrolaynen, Valentin Perminov, and Alexander Pergament**

Petrozavodsk State University, Petrozavodsk, 185910, Russia

\* Corresponding author: e-mail velichko@petrsu.ru

## Abstract

We explore a prototype of an oscillatory neural network (ONN) based on vanadium dioxide switching devices. The model system under study represents two oscillators based on thermally coupled VO$_2$ switches. Numerical simulation shows that the effective action radius $R_{TC}$ of coupling depends both on the total energy released during switching and on the average power. It is experimentally and numerically proved that the temperature change $\Delta T$ commences almost synchronously with the released power peak and T-coupling reveals itself up to a frequency of about 10 kHz. For the studied switching structure configuration, the $R_{TC}$ value varies over a wide range from 4 to 45 μm, depending on the external circuit capacitance $C$ and resistance $R_i$, but the variation of $R_i$ is more promising from the practical viewpoint. In the case of a "weak" coupling, synchronization is accompanied by attraction effect and decrease of the main spectra harmonics width. In the case of a "strong" coupling, the number of effects increases, synchronization can occur on subharmonics resulting in multilevel stable synchronization of two oscillators. An advanced algorithm for synchronization efficiency and subharmonic ratio calculation is proposed. It is shown that of the two oscillators the leading one is that with a higher main frequency, and, in addition, the frequency stabilization effect is observed. Also, in the case of a strong thermal coupling, the limit of the supply current parameters, for which the oscillations exist, expands by ~ 10 %. The obtained results have a universal character and open up a new kind of coupling in ONNs, namely, T-coupling, which allows for easy transition from 2D to 3D integration. The effect of subharmonic synchronization hold promise for application in classification and pattern recognition.

## 1. Introduction

An artificial neural network is one of the most promising approaches for the development of the next generation of computing architectures realizing the brain-inspired massively parallel computing paradigm [1]. The latest results of neurophysiological experiments proved the significant role of synchronous oscillations of neural activity in the information processing [2]. Therefore, the research dynamics of oscillatory neural networks (ONN) and their application to solve the problems of classification, clusterization and pattern recognition attracts a lot of attention. [3-5]. Hardware implementations of oscillatory neural networks are based both on current CMOS devices (e.g., phase-locked loop circuits [6] or Van der Pol oscillators [7]) and on emerging new devices, such as, spin-torque nano-oscillators [8], switches based on materials with metal-insulator [1, 9] or charge density wave [10, 11] transitions, and oxide RRAM [12, 13].

In an ONN, an elementary cell comprises an oscillator circuit, and the cells are locally coupled by resistors or capacitors [9-13]. Also, in the work [7], the variable resistive coupling via a memristive device has been proposed. Factually, such a type of coupling might be controllable, since the transition of the memristor between the ON and OFF states would bring about the coupling strength hopping between the strong-coupling and weak-coupling modes.

In this paper, we would like to consider the opportunity to connect oscillators via thermal coupling due to heat spreading through the substrate where the oscillators' elements are placed (Fig.1). To provide mutual coupling the circuit elements should act as the heat source of variable power while generating oscillations (any resistive element conducting current will do) and at the same time circuit parameters should strongly depend on temperature. Here we could mention oscillator schemes with oxide switches that have S-shape I-V characteristic and are based on such materials as $VO_x$, $NbO_x$, $TiO_x$, $TaO_x$ [4, 14-16]. The switching mechanism could be caused either by metal-insulator phase transition (MIT) [4, 14] or by formation / destruction of a channel because of ion migration mechanism [15, 16]. A channel characterized by metal-type conductivity is often formed during transition to low impedance line of I-V curve, and while the parameters of the channel have low dependence on temperature, the current there is rather high and causes local heating of the substrate. If the structure is in high impedance state the channel is characterized by semiconducting type of conductivity and its parameters depend heavily on the temperature. Such alternative switching from ON to OFF states preconditions mutual thermal coupling of two oscillators. $VO_2$-based switch is a prominent representative and model object that realizes S-shape I-V curve with parameters depending on the temperature of the environment.

The $VO_2$-based oscillators may be linked to each other by a thermal coupling taking into account the above-discussed heat-induced switching mechanism. Note that in this case, the coupling strength could also be made controllable via the variation of the heating intensity.

Vanadium dioxide undergoing a MIT is currently considered as one of the key materials for neuromorphic oxide electronics [1, 9]. Two-terminal thin-film metal/oxide/metal devices based on $VO_2$ exhibit S-type switching, and in an electrical circuit containing such a switching device, relaxation oscillations are observed under certain conditions, namely, when the load line intersects the I-V curve at a unique point in the negative differential resistance (NDR) region. The switching effect in vanadium dioxide is associated with the MIT occurring in this material at $T_t = 340$ K [17-19]. The switching effect in relatively low electric fields (namely, if the threshold switching field does not reach a value at which the development of high-field effects becomes feasible [19]) is purely thermal, i.e. it is governed by the current-induced Joule heating of the oxide film between two metal electrodes to the transition temperature $T_t$.

In our previous study [20], the possibility of using the thermal coupling to control the dynamics of operation of $VO_2$ oscillators has been demonstrated based on a model experiment with a 'VO$_2$ switch-microheater' pair. We have explored the synchronization and desynchronization modes of a single oscillator with respect to an external harmonic heat impact. The time constant of the temperature effect for the considered system configuration is $\tau \sim 150$ μs, which allows working in the oscillation frequency range of up to 10 kHz. The minimum temperature sensitivity of the switch is $\delta T_{switch} \sim 0.2$ K, and the effective action radius $R_{TC}$ of the switch-microheater thermal coupling has been estimated to be not less than 25 μm [20].

In the present study, we examine the behavior of two thermally coupled $VO_2$-based oscillators (Fig. 1). The work is focused on the following main problems: i) to identify the peculiarities of the mutual influence of two switches on the shape of their oscillation spectra and on synchronization effect on subharmonics; ii) to establish the possibility of controlling the magnitude of the effective interaction radius via the circuit parameters ($R$, $C$, $I_D$), and iii) to determine the frequency range of the coupling existence.

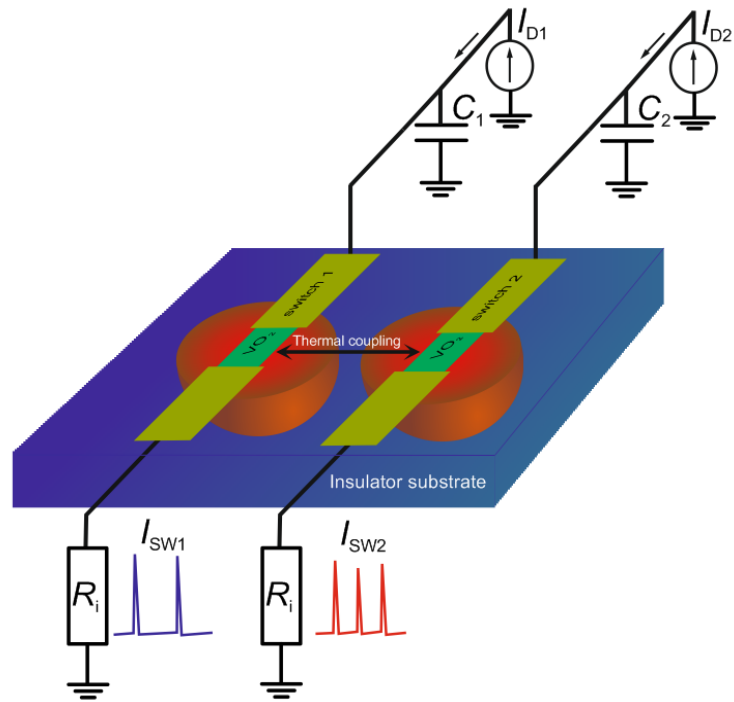

**Figure 1** [Color online] Schematic diagram of two thermally coupled oscillators.

## 2. Experimental procedure

Vanadium oxide films were obtained by DC magnetron sputtering of a 2" metal vanadium target using an AJA Orion 5 sputtering system. The residual vapor pressure did not exceed $10^{-6}$ Torr. R-cut sapphire was used as a substrate, possessing a comparatively high thermal conductivity $\chi$ ~35 W/(m·K) and insulating properties. This allowed one to place two switches in close proximity to each other, ensuring their galvanic isolation and effective temperature interaction. Prior to the deposition process, substrates were preliminarily cleaned in RF argon plasma for 5 minutes at a source power of 50 W and gas pressure of 3 mTorr.

Reactive sputtering was carried out at room temperature in a mixture of Ar and $O_2$ gases at pressure of 5 mTorr and an inlet ratio of 14 sccm and 2 sccm, respectively. During the deposition, the power of the DC discharge was ~ 200 W, the target–substrate distance was ~ 12 cm, and the sputtering time was 20 min. The thickness of the resulting vanadium oxide thin films was ~ 250 nm [20]. Next, electrically isolated planar microstructures were formed at distances $d$ = 12, 15, 18, and 21 μm from each other, with the shape of the contacts shown in Fig. 2a. The length $l$ and gap $h$ of the switch interelectrode space were ~ 3-4 μm and 2.5 μm, respectively.

To form the structures, two lithography stages were carried out using a μPG-101 laser lithograph. At the first stage, the vanadium oxide film regions were formed by etching in 4N $HNO_3$ through a resist mask. At the second stage, two-layer V-Au contacts to the switches were formed by means of lift-off lithography with the thicknesses of the V and Au layers of 20 nm and 40 nm, respectively. After lithography, the structures were annealed in air at temperature of 380 °C for 10 min. The X-ray structural analysis showed that annealing was accompanied by partial oxidation and crystallization of the vanadium oxide films with formation of $V_2O_5$, $V_2O_3$ and, predominantly, $VO_2$ phases. This was also confirmed by the temperature dependence of the film conductivity (Fig. 2b) measured by the four-probe method, which demonstrated a conductivity jump of ~$10^2$ at the transition temperature, $T_t$ ~320 K. The temperature coefficient of resistance (TCR) at room temperature, calculated from the graph of Fig. 2b, was ~2.1 %/K.

It is known that the values of $T_t$ and conductivity jump in $VO_2$ films can vary in a wide range depending on sputtering conditions and film thickness [21]. Optimization of these values for switching structures and time-stable oscillations is a separate task. Nevertheless, switches based on our $VO_2$ films had a pronounced S-shape I-V curve and demonstrated stable oscillations.

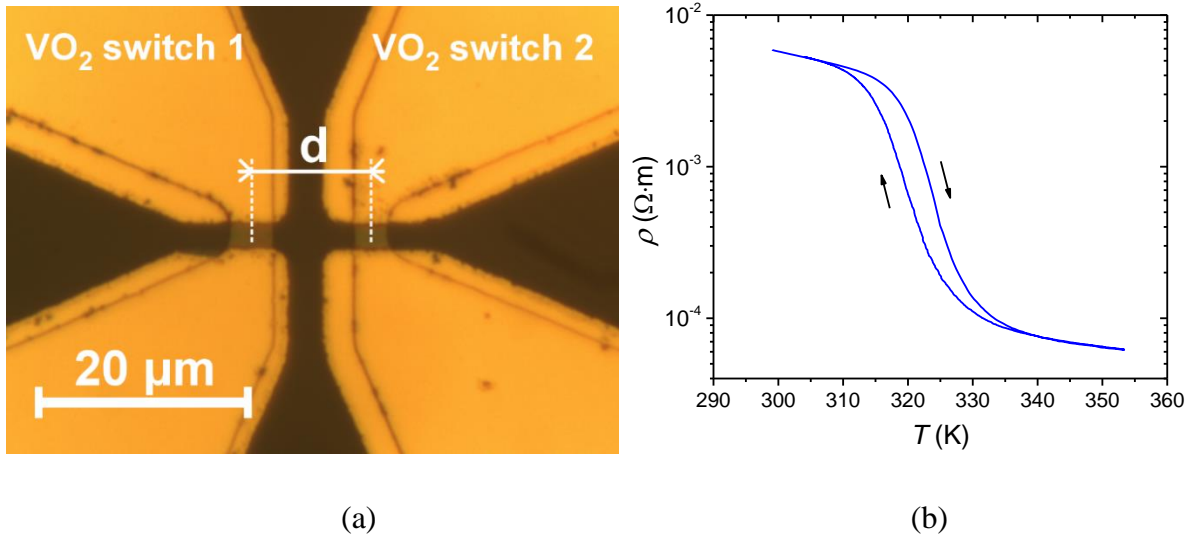

(a)　　　　　　　　　　　　　　(b)

**Figure 2** [Color online] (a) Image of the structure under study and (b) the temperature dependence of the film resistivity.

All electric measurements were made under external noise screening and at room temperature. Figure 3a shows the current-voltage characteristics of two adjacent microstructures that are S-shaped with the NDR regions. The occurrence of negative resistance is associated with the electrical switching effect which, in turn, arises when the VO$_2$ channel temperature reaches the MIT temperature $T_t$ [19]; at this point, the switch undergoes a sharp and reversible transition from a high-resistance state (HRS) to a low-resistance state (LRS). The channel temperature increase is due to the Joule heating by the flowing current; this causes the existence of the threshold currents of switching $I_{th}$ and holding $I_h$ of the structure in the metallic state. The currents on the I-V characteristic between $I_{th} = 0.4$ mA and $I_h = 1.1$ mA limit the NDR region. The threshold voltages of switching $V_{th}$ ~5.6 V and holding $V_h$ ~2 V correspond to the currents $I_{th}$ and $I_h$, respectively. Average values of the HRS and LRS resistances determined from the I-V characteristics of the structures are $R_{OFF}$ ~ 16 k$\Omega$ and $R_{ON}$ ~ 200 $\Omega$, respectively.

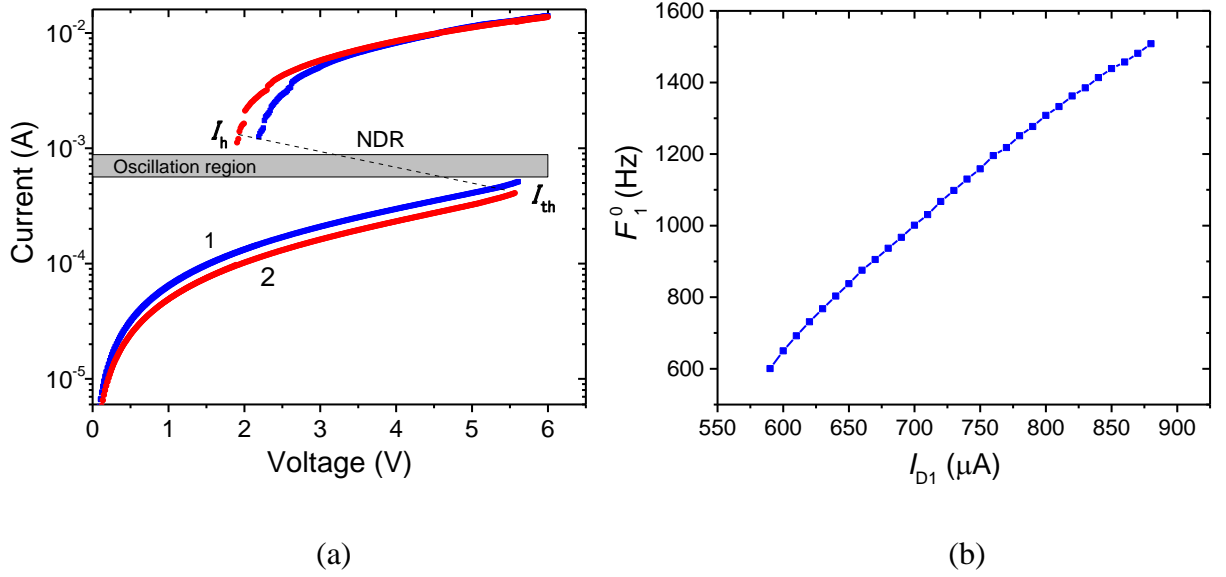

(a)                                                                (b)

**Figure 3** [Color online] (a) I-V characteristics of structures 1 and 2 measured at $R_i$=250 Ω. (b) Self-oscillation frequency $F^0_1$ as a function of supply current $I_{D1}$.

The experimental circuitry diagram for studying the dynamics of two oscillators is depicted in Fig. 1. It consists of two galvanically isolated oscillator circuits, and the only coupling element here is the thermal coupling between adjacent switches along the substrate (see Fig. 2a). External capacitances $C_1$ and $C_2$ are connected to the switches in parallel and the power of the circuits is provided by current sources ($I_{D1}$, $I_{D2}$). If the operating point of the circuit lies within the NDR region ($I_{th} < I_D < I_h$), the switch voltage becomes unstable and self-oscillation occurs [9, 20]. The values of the currents $I_D$ and capacitances $C$ determine the natural frequencies $F^0_1$ and $F^0_2$ of each oscillator.

To limit the maximum values of the capacitance discharging current through the switch, a resistor $R_i = 250$ Ω was connected in series with it. As the current source, a two-channel sourcemeter Keithley 2636A was used. The studies of oscillation dynamics were carried out with a four-channel oscilloscope Picoscope 5442B whose maximum sampling rate was 125 MS/sec in 14-bit mode. Signal spectrum was calculated on the basis of oscillograms using fast Fourier transform (FFT) method.

## 3. Results and discussion

First, we have studied the natural oscillation frequency $F^0_1$ of a single oscillator (Oscillator 1) as a function of the $I_{D1}$ current magnitude, while Oscillator 2 is turned off. The current $I_{D1}$ is varied in the NDR range, from $I_h$ to $I_{th}$ (0.4 - 1 mA), with the steps of 10 μA, though the stable self-oscillation generation is observed not in the entire range, but only within the $I_{D1}$ values from about 590 to 880 μA. This is due to the fact that, as the operating point approaches the I-V curve thresholds ($I_{D1} \rightarrow I_h$, $I_{th}$), the internal noise [22] starts to play an important role and the oscillations transit from a steady state to a burst mode [23]. The dependence of the oscillation frequency $F^0_1$ on the supply current $I_{D1}$ is practically linear (Fig. 3b), which is obviously a consequence of the inversely proportional dependence of the charging time of the capacitor on the charging current. It is straightforward to show that

$$F^0_1 = \frac{1}{T^0_1} \approx \frac{I_{D1}}{C_1 \cdot V_{th}}. \qquad (1)$$

Expression (1) is valid provided that the capacitor charging time is much longer than the discharging time, which is true for the relation $R_i + R_{ON} << R_{OFF}$ satisfied in the overwhelming majority of experiments. Because Oscillators 1 and 2 have similar capacitances and threshold switching parameters, the dependences $F^0_2(I_{D2})$ and $F^0_1(I_{D1})$ are similar. The property of mirror symmetry of two coupled oscillators presupposes independence of the presented below results on the oscillator number, i.s. effect of Oscillator 1 on Oscillator 2 will be the same as the effect of Oscillator 2 on Oscillator 1.

In order to know how the neighboring oscillators interact, it is necessary to register the temperature response in the channel area of Oscillator 2 when Oscillator 1 operates. When the parallel capacitance $C_1$ is discharged, there should be a peak of the current and released power $P_1$ leading to a local heating of the Oscillator 1 switch channel, which in turn causes a change in the neighbor switch temperature $\Delta T_2$. Figure 4a shows the time dependence of the power $P_1(t)$ at the capacitance ($C_1 = 44$ nF) discharging and the time dependence of the temperature response $\Delta T_2(t)$ calculated from the switch resistance change and its TCR for the structures located at a distance of $d$ ~12 μm from each other. One can see that the change in temperature commences almost synchronously with the power $P_1$ peak, and $\Delta T_2$ reaches a maximum $\Delta T^p_2$ with a delay of $\tau$ ~ 15 μs. In this case, it is correct to speak about the synchronism of the power and temperature peaks since the ratio of the delay τ to the oscillation period $T_1$ in the considered range of capacitances is on average ~ 2% ($\tau$ ~ 8 - 110 μs, $T_1$ ~ 100 μs - 80 ms), and the $\tau/T_1(C_1)$ dependence is decaying (Fig. 4b). This figure also shows $F^0_1(C_1)$ dependence obtained in this case. Dependence $\Delta T_2(t)$ replicates in form $P_1(t)$, albeit the temperature peak amplitude $\Delta T^p_2$ depends on many parameters, particularly, on capacitance $C_1$, current resistor $R_{i1}$, distance between switches $d$, and substrate thermal conductivity $\chi$ and heat capacity $c$. Some of these parameters ($C$, $R_i$, $F$) can somehow be changed during the oscillator operation, while others ($d$, $\chi$, $c$) are set at the structure fabrication stage. The maximum induced temperature $\Delta T^p_2$ as a function of $C_1$ capacitance value is presented in Fig. 4c. The supply current $I_{D1}$ ~600 μA and $R_i = 250$ Ω remain constant in this experiment, and capacitance $C_1$ varies from 4 to 4400 nF. It is seen that $\Delta T^p_2$ has saturation with the increasing capacitance, apparently due to the transition from dynamic to steady-state temperature distribution with an increase in the capacitor discharging duration. At a decrease in $C_1$, the value of $\Delta T^p_2$ has a strongly declining dependence on capacitance, thus allowing this effect to be used to adjust the thermal coupling strength. The dynamic heating effects account for this decaying dependence: for the same peak power $P_1(t)$, the lower the effective action time of the current signal, the lower the $\Delta T^p_2$ is, and, as is known, the capacitor discharging time in a circuit is proportional to the capacitance, $t_{ch} \sim C \cdot (R_{ON} + R_i)$.

Figure 4d shows the experimental dependences of $\Delta T^p_2$ on the distance between structures $d$ at capacitances $C_1 = 44$ nF and $1.8$ μF. The data presented demonstrate that the smaller the distance, the stronger the temperature effect is, other conditions being equal. Further, using the results of numerical modeling with the COMSOL Multiphysics software, we will show that this dependence is not linear (it is rather close to exponential) and one can estimate the thermal coupling action radius from this dependence. At a maximum experimental value of the distance, $d = 21$ μm, the thermal coupling changes the temperature of the adjacent switch by more than 0.5 K, which, as is shown below, is sufficient to influence the frequency and phase of the oscillator based on this switch.

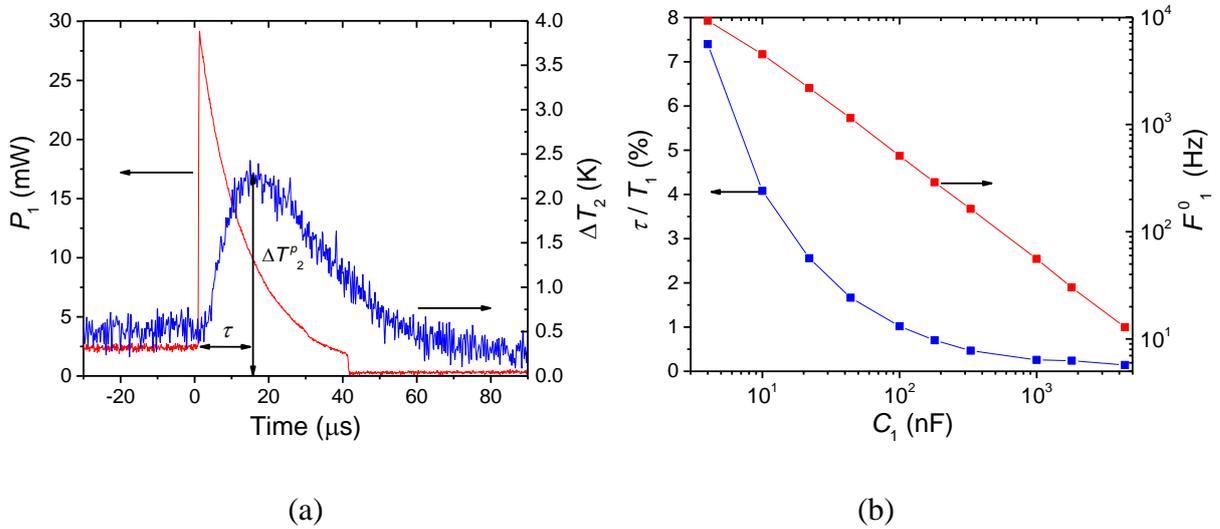

(a)                                                                    (b)

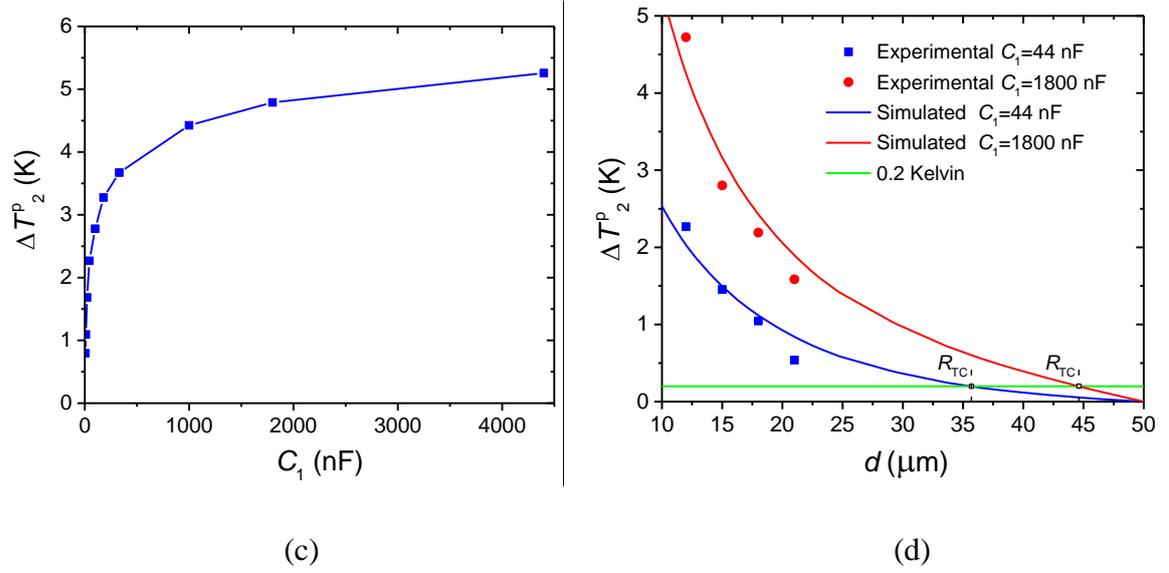

(c)                                    (d)

**Figure 4** [Color online] (a) Time dependence of temperature $\Delta T_2(t)$ in the channel area of Switch 2 at the moment of switching of Switch 1 for $C_1$= 44 nF and $d$ ~12 μm in comparison with the dependence $P_1(t)$. (b) Dependence of $\tau$ /T$_1$ and natural frequency $F^0_1$ on capacitance $C_1$. (c) Dependences of maximum temperature on capacitance and (d) on distance.

To study the dynamics of a system of two thermally coupled planar oscillators, the structures were selected at the minimum ($d = 12$ μm) and maximum ($d = 21$ μm) distances, thereby realizing conditionally "strong" and "weak" coupling, respectively. The problem consisted in recording the oscillation spectra of both oscillators when a parameter responsible for the oscillation frequency of one of them was varied. As a variable parameter, current $I_{D1}$ was changed in the range of 660-820 μA, which, in the case of no mutual thermal coupling, modified the natural oscillation frequency $F^0_1$ from 1200 to 2000 Hz. The current of the second oscillator remained constant $I_{D2} = 720$ μA and the natural frequency was $F^0_2$ ~1550 Hz. Figure 5 shows a set of paired spectra of oscillations with a step of $\Delta I_{D1} = 10$ μA.

Initially, at $I_{D1} = 660$ μA and "weak" coupling, when the first harmonic frequencies $F_1$ and $F_2$ differ significantly, the oscillators operate independently and their frequencies are close to the natural ones, i.e. $F_1 = F^0_1$ and $F_2 = F^0_2$. The spectra resemble the single oscillator spectra, possessing a broadening at the fundamental harmonic and a weak peak at the frequency of the nearby oscillator. This peak is due to the contribution of the thermal coupling at which the sporadic mutually induced switching events occur.

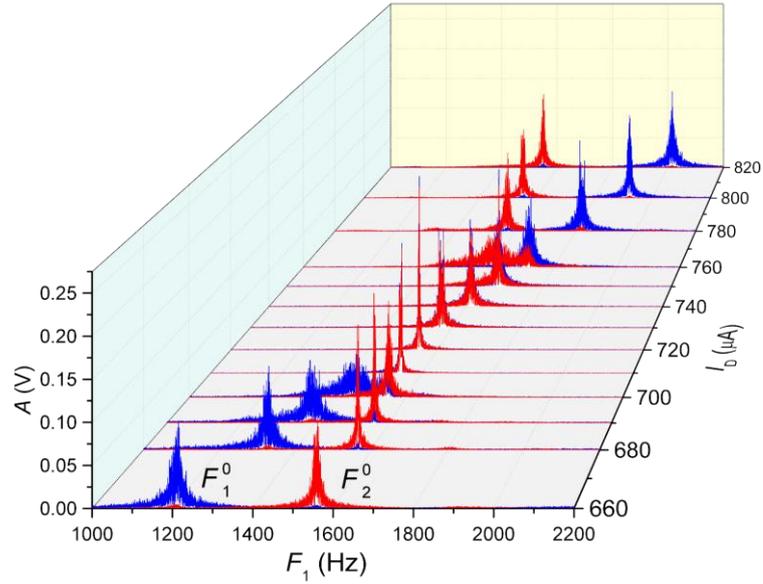

**Figure 5** [Color online] Set of paired spectra of "variable" $F_1$ (blue graphs) and "stationary" $F_2$ (red graphs) oscillators at varied supply current $I_{D1}$ of "variable" oscillator and $d = 21$ μm.

As current $I_{D1}$ increases (680-700 μA), the fundamental frequency of the first "variable oscillator" $F_1$ rises and approaches frequency $F_2$ of the second "stationary oscillator"; $F_1$ spectrum broadens and some frequency attraction effect is observed consisting in a mutual convergence of the frequencies before synchronization, when $F_1 > F^0_1$ and $F_2 < F^0_2$. In the current range 710 μA $\leq I_{D1} \leq 750$ μA, oscillators synchronize in frequency, when $F_1 = F_2$, and this effect is observed in the frequency range of 1550 to 1650 Hz. At $I_{D1} = 710$ μA, the best synchronization is observed; here, like in our work on R- and C-coupling [9], the effect of reduction in the full width half maximum (FWHM) of the first harmonic spectrum for both oscillators is noticeable, which is due to the ordering of the switching moments. Frequency synchronization is accompanied by the phase locking, and on the phase portrait, the attractor has the shape, predominantly, of an inclined straight line (Fig. 6). The two structures switch in-phase as one can see in the time phase diagram showing the phase difference between the switches (Fig. 6, insert).

The deviation of the attractor shape from an ideal straight line and the near zero noise in the phase diagram, both to positive and negative polarity, are related, in our opinion, to the above-discussed 2% time delay τ of the thermal coupling and to the random selection of the "leading oscillator". By the term "leading oscillator" we mean the oscillator that causes thermally induced forced switching of a neighboring oscillator. Hence one can surmise that, of the two oscillators, the leading one is that whose main frequency is higher, since it would earlier initiate a thermo-response on the adjacent switch. Indeed, from the spectra (Fig. 5), it is clear that before the synchronization at $F_1 < F_2$, the leading oscillator is Oscillator 2, which is confirmed by the broadening of the Oscillator 1 spectrum on the boundary with the synchronization region. For $F_1 > F_2$, the leading oscillator is Oscillator 1, and here we observe the broadening of the Oscillator 2 spectrum.

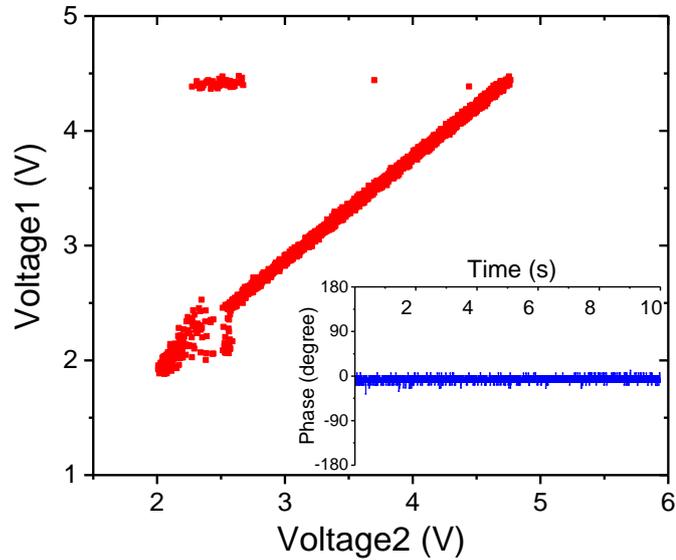

**Figure 6** Phase portrait of synchronized oscillators at a supply current of 740 μA and phase difference of the oscillators as a function of time (inset).

At $I_{D1} \geq 760$ μA, exit from the synchronization mode occurs, and as $I_{D1}$ increases, the oscillators begin to behave as uncoupled. Thus, in the weak-coupling regime, the mutual influence of the oscillators affects their oscillation spectra only if the main harmonic frequencies approach each other.

Next, we have studied the operation dynamics of a pair of coupled oscillators with a stronger thermal coupling using the structures located at a distance of $d = 12$ μm. In contrast to the weak coupling, a greater number of synchronization effects are observed in this case. Figure 7 shows the oscillation spectra when $I_{D1}$ varies in the range of 500 to 880 μA, while $I_{D2}$ remains constant and is equal to 720 μA, which corresponds to its natural frequency $F^0_2 \sim 950$ Hz. A wider frequency range is presented, as compared to that in Fig. 5, because it is necessary to analyze higher harmonics, not only the first one. To simplify the presentation, the $I_{D1}$ axis is divided into six sectors.

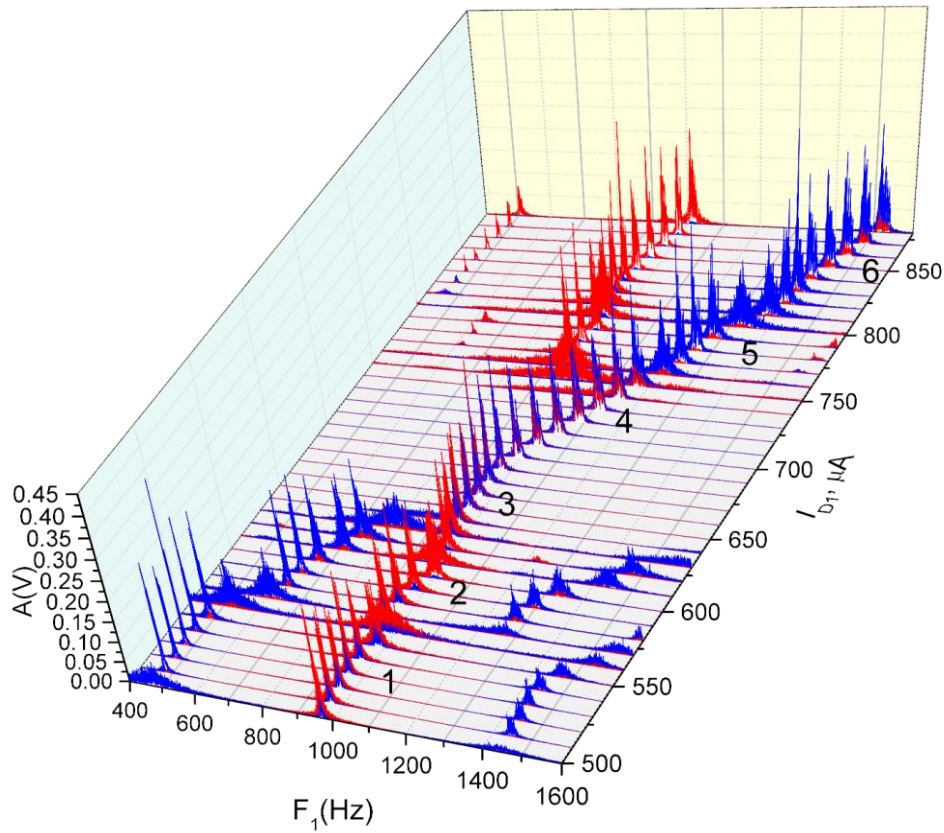

**Figure 7** [Color online] Set of paired spectra of "variable" (blue graphs) and "stationary" (red graphs) oscillators at varied supply current of "variable" oscillator for $d = 12$ μm.

In Sector 1, synchronization and frequency stabilization effects are observed; here, the period of synchronous switching $T_s$ corresponds to one period $T_1$ or two periods $T_2$, and the signals remain to be in-phase (Fig. 8a, upper panel). The frequencies differ twofold, $2 \cdot F_1 = F_2 \approx 950$ Hz, which indicates synchronization of the "stationary oscillator" first harmonic and "variable oscillator" second harmonic. Note, that the lower current limit of the "variable oscillator" operation in a single mode (without thermal coupling) is 590 μA, below which the switch is in the OFF (HRS semiconductor) state. In case of a strong thermal coupling, this boundary reduces significantly (down to $I_{D1} = 500$ μA). This is apparently due to the induced heating by $\Delta T^p_1$ of Switch 1 channel, which decreases its threshold switching-on characteristics ($I_{th}$ and $V_{th}$).

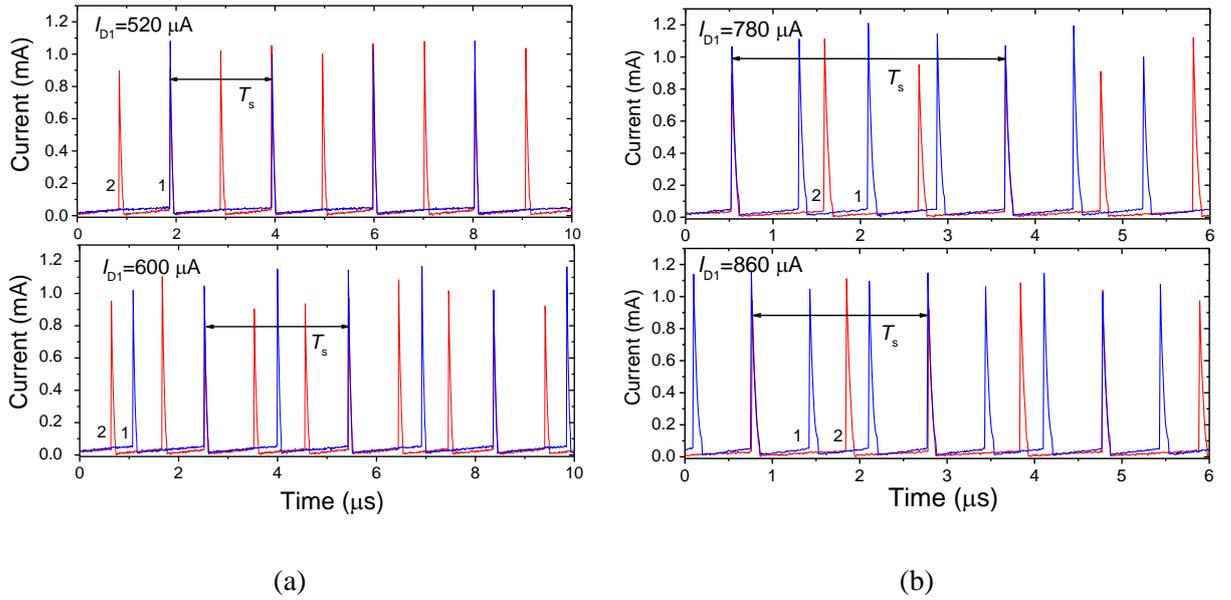

(a)                                                                    (b)

**Figure 8** [Color online] Current waveforms of partly coupled oscillators at supply currents of (a) 520 µA and 600 µA, and (b) 780 µA and 860 µA, for "variable" (blue color) and "stationary" (red color) oscillators 1 and 2, respectively.

In Sector 2, partial synchronization occurs, and the synchronicity period value $T_s = 2 \cdot T_1 = 3 \cdot T_2$ (Fig. 8a, lower panel) leads to the frequencies ratio of $3 \cdot F_1 = 2 \cdot F_2 \approx 1950$ Hz. This indicates synchronization of the "stationary" oscillator second harmonic and the "variable oscillator" third harmonic.

In Sectors 3 and 4, complete frequency and phase synchronization at the fundamental harmonics is observed; this region is much wider than in the weak coupling case. Sector 3 is characterized by the frequency stabilization effect. The change in $I_{D1}$ does not lead to a change in $F_1$ or $F_2$ value, $F_1 = F_2 \approx 1000$ Hz. Frequency stabilization occurs because the leading oscillator here is a "stationary" oscillator in the presence of a strong coupling; this effect is characteristic of Sector 1 too.

In Sector 5, the synchronicity period value $T_s = 4 \cdot T_1 = 3 \cdot T_2$ (Fig. 8b, upper panel) leads to the frequency ratio $3 \cdot F_1 = 4 \cdot F_2 \approx 3750$ Hz. This indicates synchronization of the "stationary oscillator" fourth harmonic and the "variable" oscillator third harmonic.

Finally, in Sector 6, the synchronicity period value $T_s = 3 \cdot T_1 = 2 \cdot T_2$ (Fig. 8b, lower panel) leads to the frequency ratio of $2 \cdot F_1 = 3 \cdot F_2 \approx 3000$ Hz. This indicates synchronization of the "stationary" oscillator second harmonic and the "variable" oscillator third harmonic, which is a mirror image of the synchronization type in Sector 2.

At the transition between synchronization regions, the spectrum broadening of the "driven oscillator" is observed, and the signal spectrum is similar to the spectrum of a chaotic signal. The largest shift of the stationary oscillator fundamental frequency $F_2$ occurs when the main harmonics are synchronized (Sector 4). At synchronization of higher harmonics in Sectors 2 and 4-6, we also observe the effect of a shift of the main harmonic $F_2$, yet it is much weaker. Here we also observe the effect of FWHM reduction of the fundamental harmonics at synchronization which is similar to that in case of weak coupling. This is a universal phenomenon and it correlates with other studies where decrease of phase noise of coupled oscillators is shown [24]. Nevertheless, the issue of FWHM reduction range dependence on synchronous subharmonics ratio remains open and could be the object of a separate research.

We observe synchronization effect on subharmonics in all six sections and the equality $k_1 \cdot F_1 = k_2 \cdot F_2 = F_s$, is realized, where $k_1$ and $k_2$ – numbers of subharmonics synchronizing at frequency $F_s$ (subharmonic synchronization frequency). Thus, it is possible to introduce the definition of subharmonic ratio (*SHR*):

$$SHR = \frac{k_1}{k_2} = \frac{F_2}{F_1} = \frac{T_1}{T_2}.$$  (2)

Due to the fact that $k_1$ and $k_2$ are integers, the values of *SHR* are discrete which allows for the conclusion about multi-level synchronization. Undetermined modes are observed during transition from one discrete level of synchronization to another one; therefore, we may introduce a concept of synchronization efficiency $\eta$, and provide its computation algorithm below. Values of *SHR* together with $F_s$ and $\eta$ are the basic parameters describing the effect of subharmonic synchronization.

We have introduced the following computation algorithm to determine and estimate synchronization efficiency $\eta$ and subharmonic ratio *SHR*. By numerical analysis of oscillograms of 10 s duration (Fig. 8 a,b,) we found some moments when two peaks were synchronized (with accuracy $\sim 10^{-5}$ s) and determined period $T_s$, then we calculated how many $M$ periods of each oscillator fall between them (let us denote these values $M^1$ and $M^2$). Then we calculated the number of repetitions $N$ with the same $M$ for each oscillogram (denote them $N^1$ and $N^2$), Table 1 shows their values for $I_{D1} = 610$ µA and $I_{D1} = 780$ µA. The values $M^1_{max}$ and $M^2_{max}$ corresponding to maximum $N$ (denote them as $N^1_{max}$ and $N^2_{max}$) were selected from the table. It is obvious (see formula 2) that their ratio corresponds to the value *SHR*:

$$SHR = \frac{T_2}{T_1} = \frac{M^1_{max}}{M^2_{max}}.$$  (3)

Due to the fact that in the general case $N^1_{max} \neq N^2_{max}$, the efficiency of synchronization $\eta$ was determined as percentage ratio of the least value of $(N^1_{max}, N^2_{max})$ to the total number of synchronous periods:

$$\eta = \left(\frac{\min(N^1_{max}, N^2_{max})}{\sum N}\right) \cdot 100\% . \qquad (4)$$

At the final step we compared the values of $\eta$ with the limit of 90% that we have selected. Oscillations were considered synchronized at $\eta > 90\%$, if not, they were considered desynchronized (chaotic) ones.

When analyzing the example of $SHR$ and $\eta$ calculation for two different values $I_{D1}$ (see Table 1) we may come to the conclusion that we obtain $M^1_{max}=3$, $M^2_{max}=4$, for current $I_{D1} = 610$ µA which corresponds to $SHR=3/4$, however, calculation $\eta \sim 72\%$ does not match the limit of 90%, therefore we consider the oscillations as desynchronized ones. For $I_{D1} = 780$ µA we obtain $M^1_{max}=4$, $M^2_{max}=3$, $SHR=4/3$, $\eta \sim 99\%$, respectively and the oscillations are considered as synchronized.

It should be noted that this algorithm is more efficient and fast than, for instance, analysis of oscillation spectrum, as $SHR$ values may be calculated according to the spectrum analysis only in some specific cases. Calculation of $\eta$ values on the basis of spectrum also involves considerable difficulties. This primarily is due to the fact that our algorithm is resistant to period fluctuation (as the number of periods is taken into account but not their duration), while on a spectrum this is expressed through line broadening and with the increase of harmonics ratio $k$ its amplitude decreases. One last thing: FFT algorithm requires cumbersome arithmetic and all of this complicates analysis.

**Table 1** Results of experimental current waveforms processing according to the algorithm and calculation of synchronization efficiency $\eta$ at currents: $I_{D1}$ =610 µA ($N^1_{max}$= 1716, $N^2_{max}$=1708, $M^1_{max}$=3, $M^2_{max}$=4, $SHR$=3/4) and $I_{D1}$ =780 µA ($N^1_{max}$ = $N^2_{max}$=3105, $M^1_{max}$=4, $M^2_{max}$=3, $SHR$=4/3)

| | $I_{D1}$=610 µA | | $I_{D1}$=780 µA | |
|---|---|---|---|---|
| $M$ | $N^1$ | $N^2$ | $N^1$ | $N^2$ |
| 1 | 7 | 7 | 11 | 11 |
| 2 | 411 | 0 | 0 | 0 |
| 3 | **1716** | 411 | 0 | **3105** |
| 4 | 42 | **1708** | **3105** | 33 |
| 5 | 181 | 14 | 33 | 1 |
| 6 | 5 | 35 | 0 | 0 |
| 7 | 17 | 183 | 1 | 0 |
| 8 | 0 | 0 | 0 | 0 |

| | | |
|---|---|---|
| $\eta$, % | **71,79** | **98,57** |

The result of *SHR* calculation according to the algorithm (2-4) plotted versus $I_{D1}$ is shown graphically in Fig.9. Here we observe six discrete *SHR* states, which correspond to the results of spectrum analysis given above (Fig. 7). However, numerical algorithm allows defining the current boundaries more distinctly for each *SHR* value. If the limit for $\eta$ goes down intermediate states that were not defined previously may become visible (for instance, at 80% limit, at $I_{D1}$=620 μA, *SHR*=3/4 and $\eta$~86% thus the state is considered to be synchronous).

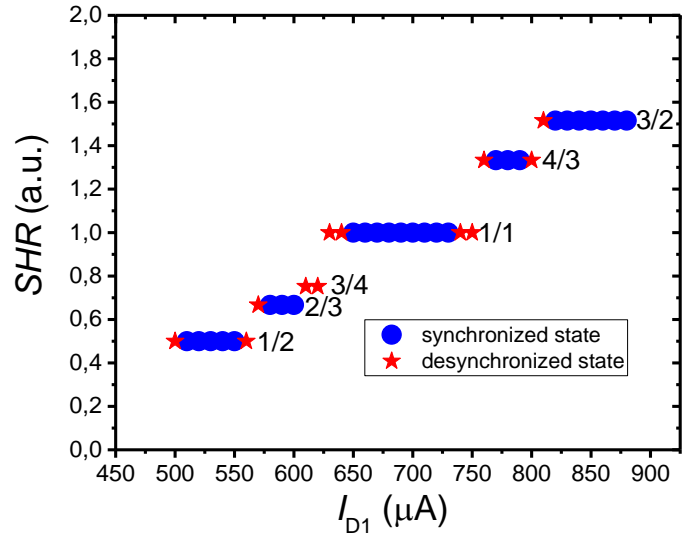

**Figure 9** [Color online] *SHR* dependence on current $I_{D1}$ of "variable" oscillator (strong thermal coupling).

If the SHR calculation algorithm is applied in the case of weak coupling (See Fig.5), then synchronization on the main harmonics (*SHR*=1/1) is observed only for the current range $I_{D1}$=710-740 μA, and there is no synchronization on subharmonics.

To study the effect of locality of the thermal coupling, one have to find out how the effective oscillators' interaction radius $R_{TC}$ varies with variations in the values of capacitance $C$ and current resistance $R_i$ in the circuit. Here, it is to be clarified that by $R_{TC}$ we mean an average radius at which the peak temperature change $\Delta T^p$, caused by electrical switching when the capacitor is discharged, reaches a quite determined value (Fig. 10a). In the experimental part, we have pointed out that $\Delta T^p$, and hence $R_{TC}$, can depend both on the circuit parameters ($C$, $R_i$), and on the switch dimensions and substrate thermal constants ($d$, $\chi$, $c$). In this work, we have used the invariable switch dimensions settable in experiment, though the study of scaling issues might be the subject of an additional research. From general considerations, it is clear that the supply current $I_D$ does not affect $R_{TC}$, because it determines only the period of oscillations, see formula (1). Therefore, the above presented experimental dependences of the oscillation spectra on $I_D$ are a consequence of the change in the frequencies of the first harmonics at constant $R_{TC}$. As in [20], the value of $R_{TC}$ is determined as the distance at which $\Delta T^p$ change is not less than 0.2 K.

For the numerical simulation, a complex oscillator model has been built using COMSOL Multiphysics software. The model consists of a physical model of the switching region (Fig. 10a) and an external electrical circuit similar to that shown in Fig. 1. The physical model presents the $100 \times 100 \times 100$ μm sapphire substrate and the switch consisting of a VO₂ film in the form of a 250 nm thick and 4 μm wide strip and gold contacts placed over it with a gap of 2.5 μm.

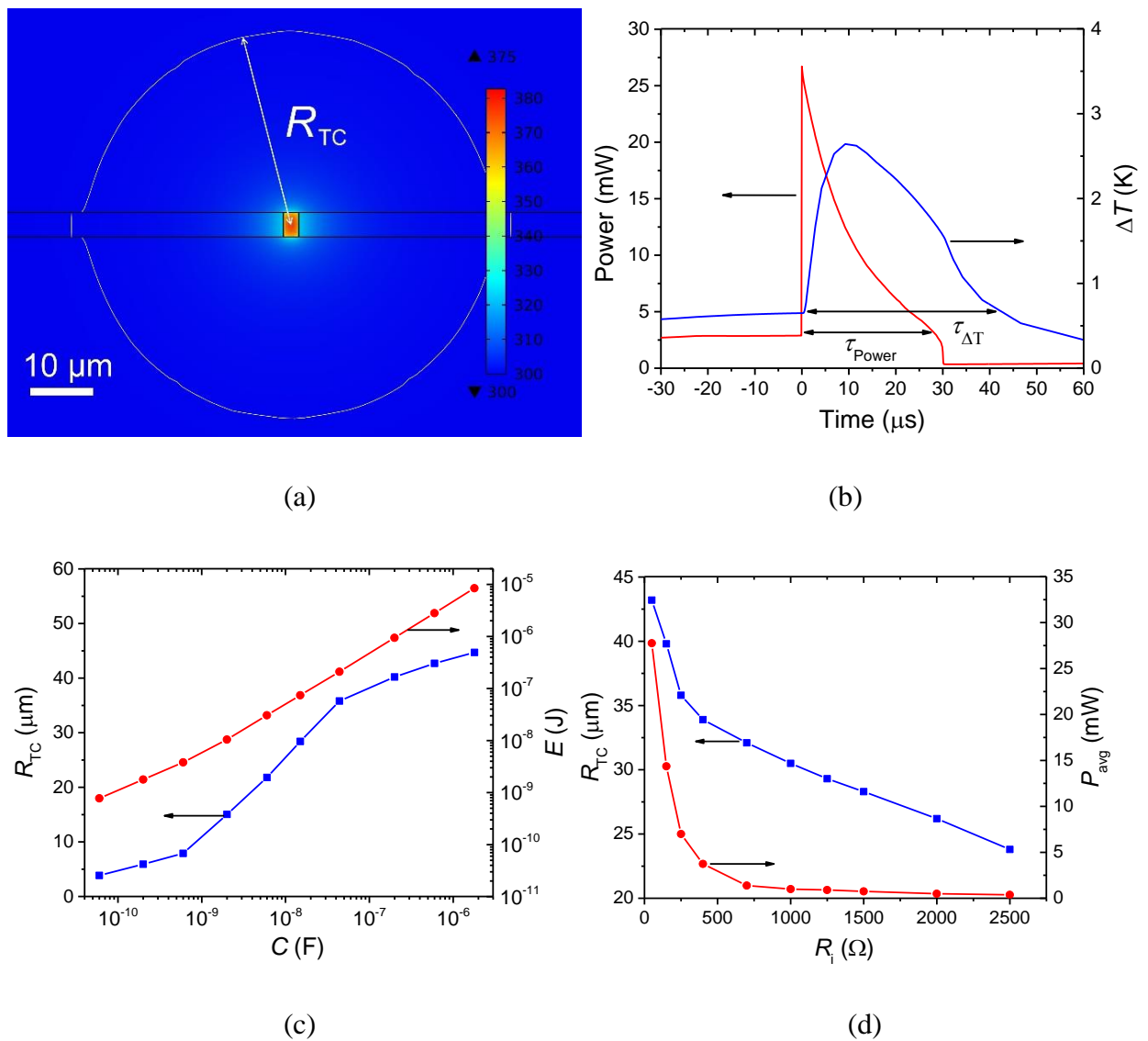

(a)                                                                    (b)

(c)                                                                    (d)

**Figure 10** [Color online] Results of simulation of: (a) temperature distribution in the switch region and shape of the thermal interaction effective radius area $R_{TC}$; (b) time dependence of the power

released and the temperature change at the moment of switching for $d = 12$ μm; (c) dependence of $R_{TC}$ and $E$ on the capacitance $C$ in the oscillator circuit; (d) dependence of $R_{TC}$ and $P_{avg}$ on the current resistance $R_i$.

The values of the material constants were taken from [25], and the resistivity vs. temperature plot corresponded to the direct (i.e., when heating) branch of the dependence shown in Fig. 2b. The switching effect in this model arose by a natural way due to the thermal mechanism at the Joule heating of the channel by the flowing current. The simulation results describe the experimental data fairly well. For example, the model I-V curve showed 90-% matching of the threshold parameters ($R_{ON}$, $R_{OFF}$, $V_{th}$, $V_h$) with their experimental values, and its form corresponded to the experimental dependence of Fig. 3a.

When modeling the time-dependent behavior of the electric circuit, stable oscillations have been obtained; supply current has been set $I_D = 600$ μA, current resistor $R_i = 250$ Ω, and capacitance $C = 44$ nF. Figure 10b shows the results of modeling the power time dependence at the capacitor discharging and the corresponding temperature change at a distance of 12 μm from the switch. The maximum temperature change $\Delta T^p \sim 2.5$ K and the delay time $\tau \sim 12$ μs in the model fit satisfactorily to the experimental data (Fig. 4a). The capacitor discharge duration $\tau_{Power} = 30$ μs and the thermal response time $\tau_{\Delta T} = 40$ μs are 1.5 times higher than the corresponding experimental values, which is quite acceptable for the proposed model and is a good approximation. Thus, we arrive at a conclusion that the experimental and model results are in accordance, which allows us to estimate the value of $R_{TC}$ depending on the circuit parameters.

Figure 4d shows the results of modeling the experimental dependences $\Delta T^p{}_2(d)$ measured at two different capacitances $C_1$ and at $R_i = 250$ Ω. These dependences are exponentially decaying and they describe the experiment well. The distance $d$, at which the curves intersect the $\Delta T^p{}_2 = 0.2$ K line, may serve as an estimate of the effective action radius $R_{TC}$ for thermal coupling.

The results are presented in Figs. 10c-d. As the capacitance $C$ increases, the effective radius of $R_{TC}$ increases (Fig. 10c), and as the current resistance $R_i$ increases, on the contrary, it falls (Fig. 10d). To clarify the physics of these regularities, we assume that the increase in capacitance is accompanied by an increase in the stored energy, whose part $E$ is released during switching and leads to the channel Joule heating. The released energy $E$ can be estimated from the formula:

$$E = \left(\frac{C \cdot V_{th}^2}{2} - \frac{C \cdot V_h^2}{2}\right) \cdot \frac{R_{ON}}{R_{ON} + R_i}. \tag{5}$$

In this expression, we take into account the fact that the capacitance is discharged within the threshold voltages $V_{th}$ and $V_h$, and the current resistor, connected in series with the switching structure, accepts some of the discharge energy. The corresponding graph of $E(C)$ is shown in Fig. 10c. We have also calculated the average power of the current pulse $P_{avg}$, which is equal to the ratio of energy $E$ to the capacitor discharge duration $\tau_{Power}$ (Fig. 10b). In the model experiment, the current resistor value is constant $R_i = 250\ \Omega$, and capacitance $C$ varies from 60 pF to 1.8 μF. It can be seen from Fig. 10c that the value of $E$ varies by almost 5 orders of magnitude, from $7.8 \cdot 10^{-10}$ J to $8.5 \cdot 10^{-6}$ J, and the average power, as calculation shows, remains almost invariable, $P_{avg} \sim 8$ mW. This leads to a change in the effective interaction radius in the range of 4 to 45 μm, and the dependence of the change in $R_{TC}$ on $\lg(C)$ resembles a linear one. Thus, by varying $C$, one can change the $R_{TC}$ value, though this approach has a drawback, since alongside the RTC, the natural oscillation frequency $F^0$ changes too (Fig. 4b), and as is known, frequency plays a crucial role in the oscillatory network operation. In practice, this is of importance for pattern recognition by the frequency shift keying method [26].

Modeling of the $R_{TC}(R_i)$ dependence has been carried out at constant capacitance $C = 44$ nF and $R_i$ varied in the range from 50 Ω to 2.5 kΩ (Fig. 10d). The simulation results have revealed an interesting fact: an increase in $R_i$ leads to an increase in the structure ON state resistance, whose value varies from 170 Ω to 1 kΩ. This is due to the effect of incomplete heating of the $VO_2$ channel by a current (power) pulse at the temperature front propagation from the center to periphery, which results in a situation where the central part of the channel is in the metal phase, while the rest part is in the insulator phase, and their ratio decreases with increasing $R_i$. When $R_i$ is varied, energy $E$ released by the capacitor is practically unchanged and is of the order of $1-2 \cdot 10^{-7}$ J, and $P_{avg}$ varies in the range of 27.7 to 0.4 mW. As the $R_i$ value increases, the $R_{TC}$ radius decreases from 43.2 to 23.8 μm, that appears to be linked to the $P_{avg}$ decrease, see Fig. 10d.

Thus, the obtained model data indicate the existence of mediated dependences of $R_{TC}(E)$ and $R_{TC}(P_{avg})$, which is revealed by the separate varying of $C$ and $R_i$; in this case, $R_{TC}$ value can vary in a wide range, from 4 to 45 μm, depending on $C$ and $R_i$ parameters. That is, by changing $C$ and $R_i$, it is possible to regulate the local interaction region in a system of thermally coupled oscillators. Meanwhile, we believe that variation of $R_i$ is more promising from the practical viewpoint, inasmuch as it brings about no significant changes in the natural frequency.

Figure 10a shows the channel temperature distribution at the time moment when the temperature reaches peak changes during the oscillation process. One can see that $R_{TC}$ radius describes a practically regular circle around the switch perimeter and, obviously, describes a semisphere through the substrate thickness (we recall that $R_{TC}$ corresponds to $\Delta T^p \sim 0.2$ K). The maximum temperature in the channel center reaches 380 K, which is slightly higher than $T_t$. In our subsequent publications, we are going to present the results and effects related to the dynamics of temperature changes during the channel formation.

In conclusion of this section, we would like to comment on the following issue. One cannot but admit that the thermal coupling in an electronic circuit looks rather unusual and it may even seem unreliable. It should be noted however that various devices to control the heat flows, similar to their electronic counterparts to control the charge carrier flows, have lately been extensively discussed in the literature [27-33]. These studies have resulted in a distinct research area termed as "phononics" [29] or "thermotronics" [28]. Thermal transistors [28-30, 32, 33], thermal diodes [29], thermal memory cells [28, 29, 32], and a lot of other devices have been proposed. Interestingly, some of these devices are based on vanadium dioxide [27, 28, 31, 32]. Thus, the approach developed in the present study just follows this trend and, furthermore, combines thermotronics with electronics to create an ONN.

### 4. Conclusions

We have studied the dynamics of thermally coupled $VO_2$-based oscillators and shown that, in the case of a "weak" coupling synchronization is accompanied by effects of attraction and reduction of the main spectrum harmonics. In the case of a "strong" coupling, the number of effects increases, synchronization can occur on subharmonics resulting in multilevel stable synchronization of two oscillators. An advanced algorithm for synchronization efficiency and subharmonic ratio calculation is proposed. Between the modes of "multiharmonic" synchronization, there are transient regimes with a chaotic behavior of oscillations and a broadened spectrum. It is shown that, of the two oscillators, the leading one is that with a higher main frequency, and the frequency stabilization effect occurs because the leading oscillator is a "stationary oscillator". Also, in the case of a strong thermal coupling, the limit of the current parameters $I_D$, for which the oscillations exist, expands by $\sim 10$ % (from $I_{D1} = 590$ μA to $I_{D1} = 500$ μA).

The effect of the thermal coupling of two oscillators, which can be considered as a new type of coupling to be used in an oscillatory neural network, is shown in the present paper for the first time. The presented method of interaction by heat exchange allows realization of the oscillator coupling without the use of electrical elements (resistors, capacitors, etc.), thereby simplifying the scaling problem, i.e., the problem of increasing the packing density of the oscillators per unit area. In addition, the presented method implements the "all with all" coupling topology in a natural way, within the effective coupling action radius, and allows for easy transition from 2D to 3D integration, because the thermal effect spreads through the volume surrounding the switch.

Yet another benefit is an electrical decoupling with respect to both direct and alternating current, since the communication channel used is not an electrical one. By changing the conditions for implementing the coupling method, the ONN functionality can be varied. For example, it is of interest to study the influence of one oscillator on another at significantly different switch parameters ($V_{th}$ or channel sizes) with the $R_{TC}$ values differing several times. In this case, one can expect a unidirectional influence of one oscillator on another, which might be used to separate input and output oscillatory neuro-circuits. One more pathway may be the use of current resistors with a large TCR built-in to the substrate, whose location with respect to other switches would create the switch-to-resistor coupling, complimentarily to the switch-to-switch coupling. Such a diversity of couplings would result in a more flexible mechanism for the ONN designing based on T-coupling effect.

Interestingly, T-coupling turns out to be a very fast mechanism for effect transferring, according to its physical principle, which accounts for the synchronicity of the $\Delta T$ response with respect to the peak power $P(t)$ (Figs. 4a and 10b). The fact is that the phonon propagation time to a distance of 6 μm does not exceed 1 ns for sapphire, given that the material sound speed is about $10^6$ cm/s [34]. It has been shown experimentally that there is a distinguishable thermo-response up to a frequency of 10 kHz (Fig. 4b), and this is surely not the limit. As we have already shown by numerical modeling [20], thermal coupling may be also realized in GHz range when dimensions of a switch decrease to nanometric dimensions. Practically, oscillations on $VO_2$ structures are observed at frequencies up to MHz [35], thus meaning that thermal mechanisms work at these frequencies    Also, the power and temperature peaks are almost synchronized, since the ratio of the delay τ to the oscillation period $T_1$, in the considered range of capacitances, is on average less than 2%.

We have shown that an effective change in the $R_{TC}$ value can be achieved by varying the external circuit parameters $C$ and $R_i$, and this value varies from several to tens of microns. This allows one to greatly change the area (or volume, in case of 3 D integration) of the coupling region in which tens and hundreds of switches can be placed. The main physical process ensuring the thermal coupling is the Joule heating of the switching region. As the simulation shows, $R_{TC}$ depends on both the total released energy $E$ and the average input power $P_{avg}$, so future research could be focused on an attempt to derive an empirical $R_{TC}(E, P_{avg})$ relation.

Nevertheless, using the dependences obtained in this work (Fig. 10), it is possible to estimate the limits of miniaturization of R- and C-coupled oscillatory networks [9], since in this case the thermal coupling is a parasitic effect and can significantly influence the network operation modes provided that the switches are located at distances closer than $R_{TC}$.

Planar structures based on vanadium oxides can be modified under the action of electron-beam and UV irradiation [36, 37]; they can be fabricated using the low-temperature sol-gel technology [38, 39] and vanadium-oxide-based lithography resists [40, 41]. All this emphasizes once again the prospects of using vanadium oxides in the electronics of the future, in particular, for the intelligent ONNs designing, including those based on the phenomenon of thermal coupling described in the present paper.

The effect of subharmonic synchronization studied in this paper might increase the number of states of coupled oscillators system. Usage of this effect for pattern recognition and classification problem (similar to [3]) will allow for increase of stored pattern number in ONN while working according to FSK-scheme without increasing the number of oscillators. The authors have proposed an effective algorithm to determine the value of synchronization efficiency of two oscillators and the value of subharmonic ratio at the synchronization frequency. A notable thing is that a part of a spectrum is used efficiently with higher frequency at synchronization on subharmonics ($k_1 \cdot F_1 = k_2 \cdot F_2 = F_s$, when ($k_1$ or $k_2$)>>1) thus as if virtually increasing frequency range of an oscillator network. This effect may appear while observing transition time between two *SHR* states and requires further investigation. It is obvious that the effect of subharmonic synchronization is a fumdamental one and does not depend on the type of coupling whether it is eletrical, magnetic, optical, heating or any other one. The obtained results have a universal character and open up a new kind of coupling in ONNs, namely, T-coupling, which allows for easy transition from 2D to 3D integration, and effect of subharmonic synchronization that could be used for classification and pattern recognition.

## Acknowledgements

This work was supported by Russian Science Foundation, grant no. 16-19-00135.